\documentclass[conference]{IEEEtran}
\IEEEoverridecommandlockouts
\usepackage{cite}
\usepackage{amsmath,amssymb,amsfonts}
\usepackage{algorithmic}
\usepackage{graphicx}
\usepackage{textcomp}
\usepackage{xcolor}
\usepackage{booktabs}
\usepackage{multirow}
\usepackage{url}
\usepackage[hidelinks]{hyperref}

\def\BibTeX{{\rm B\kern-.05em{\sc i\kern-.025em b}\kern-.08em
    T\kern-.1667em\lower.7ex\hbox{E}\kern-.125emX}}

\begin{document}

\title{HeteroWear: Sensor-Aware Self-Supervised Learning
for Heterogeneous Wearable\\ Activity Recognition}

\author{
\IEEEauthorblockN{Yasaman Mohammadpour, Pouya Shaeri}
\IEEEauthorblockA{School of Computing and Augmented Intelligence \\
Arizona State University \\
Tempe, Arizona, USA \\
ymoham15@asu.edu, pshaeri@asu.edu}\author{
\IEEEauthorblockN{Yasaman Mohammadpour, Pouya Shaeri}
\IEEEauthorblockA{School of Computing and Augmented Intelligence \\
Arizona State University \\
Tempe, Arizona, USA \\
ymoham15@asu.edu, pshaeri@asu.edu}
\thanks{Extended version of a paper accepted at the 2026 IEEE International Conference on Big Data (IEEE BigData 2026). \copyright~2026 IEEE. Personal use of this material is permitted. Permission from IEEE must be obtained for all other uses, in any current or future media, including reprinting/republishing this material for advertising or promotional purposes, creating new collective works, for resale or redistribution to servers or lists, or reuse of any copyrighted component of this work in other works.}
}
}

\maketitle

\begin{abstract}
Self-supervised learning for wearable sensing is usually confined to a single dataset or device family, leaving open whether representations transfer across heterogeneous sensor configurations and whether gains arise from pretraining rather than from architecture or richer sensor input. We present HeteroWear, a sensor-aware self-supervised framework pretrained on an eight-dataset wearable corpus (173 subjects, 358{,}261 windows) spanning accelerometer, gyroscope, magnetometer, photoplethysmography, electrocardiography, electrodermal activity, respiration, temperature, and heart-rate signals across multiple body locations, devices, and native sampling rates. HeteroWear separates within-sensor temporal modeling from cross-sensor relational modeling using modality-specific patch tokenization, sensor-context embeddings, temporal and cross-sensor Transformers, and latent cross-attention fusion. Training combines structured sensor dropout, cross-configuration consistency, and per-axis masked sensor modeling; missing sensors are omitted rather than reconstructed. Under leave-one-dataset-out pretraining and subject-disjoint evaluation on PAMAP2, WISDM, HHAR, HARTH, and UCI-HAR, HeteroWear improves mean macro-F1 from 0.688 to 0.724 over the identical full-channel architecture trained from scratch, with the largest controlled gain on heterogeneous-device HHAR (0.823 versus 0.741). Relative to single-accelerometer supervised baselines, it achieves the highest macro-F1 on four of five benchmarks. Held-out-device evaluation yields a mean +0.066 macro-F1 improvement over the matched scratch model. Under 50\% test-time channel removal, HeteroWear retains 73.8\% of full-channel performance versus 64.1\% for the same architecture trained from scratch without sensor dropout. Ablations show that structured sensor dropout primarily drives robustness, while cross-configuration consistency and sensor-context encoding contribute to transfer. These results indicate that heterogeneous sensor-aware pretraining improves both cross-device generalization and resilience to changing sensor availability.
\end{abstract}


\section{Introduction}

Wearable sensing research has produced dozens of human activity recognition (HAR) datasets, each collected with a different combination of sensor modalities, on-body placements, device hardware, and sampling protocols. A model trained on chest-and-wrist accelerometry at 100~Hz from a research-grade inertial unit does not, in general, transfer to a consumer smartwatch at 32~Hz on the wrist alone, and a model that assumes a fixed input channel set breaks the moment one channel is unavailable, whether from lost connectivity, a dead battery, or a device simply not worn that day. Cross-device heterogeneity and missing-modality robustness are usually treated as separate problems, addressed either by training a bespoke model per dataset or by imputing missing channels after the fact. We ask whether a single pretrained representation, trained explicitly to expect variation in which sensors are present, can serve both purposes.

Both problems have concrete practical stakes beyond benchmark accuracy. Continuous activity and physiological monitoring is increasingly deployed across heterogeneous consumer hardware simultaneously, a research cohort might combine participants' own smartwatches with study-issued research-grade sensors, and a single participant's device mix can change mid-study as batteries die, straps break, or new hardware ships. The same pattern of mixed and changing inputs appears well beyond HAR: environmental and thermal-exposure monitoring combines low-cost modular sensing hardware~\cite{Shaeri2026} with models that draw on meteorological measurements, imagery, and physical priors~\cite{alkhaled2024webmrt,shaeri2025multimodal,shaeri2025saud,middel2026visionmrt,middel2026openmrt}, and multimodal reasoning systems must work with whichever evidence is available~\cite{beigi2025can}. A model that requires retraining, or even re-specification of its input schema, every time the sensor population shifts is a genuine deployment obstacle, independent of its accuracy on any single fixed benchmark. This motivates evaluating not only whether a representation transfers across devices, but whether it remains usable, without retraining, when the specific set of available sensors changes after deployment.

A closely related and fast-growing line of work targets the missing-modality half of this problem for physiological and wearable signals. PhysioME~\cite{lee2025physiome} combines contrastive and masked-prediction pretraining with a dedicated restoration decoder that reconstructs missing-modality tokens from observed ones. PhysioOmni~\cite{jiang2025towards} learns a decoupled multimodal tokenizer with shared and modality-private codebooks, then applies resilient fine-tuning with prototype alignment to handle arbitrary missing modalities in EEG/ECG/EOG/EMG brain-computer-interface tasks. CognifySSL v2.0~\cite{cisse2026detect} pairs contrastive and masked objectives with an explicit modality-presence predictor and reconstruction decoder for wearable IMU and physiological streams. All three share a repair philosophy: something is trained to estimate what a missing sensor would have said. HeteroWear takes the opposite design stance, common to masked-autoencoding work in vision~\cite{he2022masked,bachmann2022multimae} but less explored for cross-\emph{dataset} wearable heterogeneity: a missing sensor is never reconstructed, only omitted from the attention computation via a padding mask. This is a design difference, not a claimed superiority. These systems use different tasks and protocols, so we position HeteroWear against them conceptually (Section~\ref{sec:related}).

This distinction motivates three research questions.

\begin{itemize}
\item \textbf{RQ1:} Does heterogeneous self-supervised pretraining improve downstream transfer after controlling for architecture and sensor availability, even when the target dataset is completely excluded from pretraining?

\item \textbf{RQ2:} Does sensor-aware pretraining improve generalization to unseen devices beyond an identical full-channel architecture trained from scratch?

\item \textbf{RQ3:} Does structured pretraining improve robustness to sensor loss, and which components of HeteroWear account for this behavior?
\end{itemize}

Our contributions are: (1) a hierarchical sensor-aware architecture that separates within-sensor temporal modeling from cross-sensor relational modeling and uses latent cross-attention to fuse a variable set of available sensors without imputation; (2) self-supervised representation learning over an eight-dataset heterogeneous wearable corpus using explicit modality, location, device, and sampling-rate context together with structured sensor dropout, cross-configuration consistency, and masked sensor modeling; and (3) a controlled leave-one-dataset-out evaluation against the identical full-channel architecture trained from scratch, together with held-out-device, missing-sensor, and mechanism-ablation experiments.

\section{Related Work}
\label{sec:related}

Supervised HAR on wearable sensors has a long history of purpose-built architectures evaluated on single, fixed-configuration benchmarks. DeepConvLSTM~\cite{ordonez2016deep} combines convolutional feature extraction with recurrent layers and remains a standard strong baseline on PAMAP2~\cite{reiss2012introducing} and HHAR~\cite{stisen2015smart}. Transformer architectures~\cite{vaswani2017attention} adapted to the same single-channel setting can outperform convolutional-recurrent hybrids given sufficient data~\cite{ek2023transformer}. PatchTST~\cite{nie2022time} established patch-based tokenization as an effective general recipe for time-series Transformers; HeteroWear adopts the same patching principle but replaces PatchTST's channel-independent weight sharing with explicit context embeddings, since our channels come from different datasets, devices, and physical sensor types rather than repeated measurements of a comparable kind.

Self-supervised learning for wearable HAR largely follows the pretrain-then-fine-tune recipe from vision and language. Haresamudram et al.\ find that self-supervised pretraining improves downstream label efficiency for accelerometer-based HAR relative to training from scratch~\cite{haresamudram2021contrastive,haresamudram2022assessing}. LIMU-BERT~\cite{xu2021limu} applies BERT-style~\cite{devlin2019bert} masked reconstruction to inertial data within a small device family; we adopt the same masked-reconstruction convention for our single-channel control to test whether it alone explains our results. CroSSL~\cite{deldari2024crossl} handles missing modalities through latent masking but is only partially cross-device. Yuan et al.~\cite{yuan2024self} pretrain on 700{,}000 person-days of UK Biobank accelerometer data and show the representation generalizes across external benchmarks, demonstrating that scale can substitute for architectural heterogeneity handling when the modality is fixed to a single accelerometer stream. HeteroWear pretrains jointly across genuinely different modalities, locations, and devices at a comparatively modest data scale, testing whether architecture can substitute for scale rather than the reverse.

Structured sensor dropout is related to dropout-style stochastic regularization, including selective variants that operate on individual neurons~\cite{shaeri2025mid}; our variant removes whole sensors, so the removal is a property of the input configuration rather than of the network. Missing or heterogeneous inputs are also a general concern in multimodal learning, for example when verifying claims against text and images~\cite{beigi2025can} or estimating thermal exposure from varying combinations of imagery and meteorological inputs~\cite{shaeri2025multimodal,shaeri2025saud,middel2026visionmrt}, and learning from heterogeneous, partially labeled sources recurs in text analysis of social media~\cite{shaeri2023semi,shaeri2025sentiment,shaericlimateemote}.

\begin{table*}[t]
\caption{Qualitative comparison of closely related self-supervised methods, based on the setting each paper describes.}
\label{tab:related-comparison}
\centering
\small
\setlength{\tabcolsep}{6pt}
\begin{tabular}{lccc}
\toprule
\textbf{Method} & \textbf{Missing-modality handling} & \textbf{Cross-device} & \textbf{\# datasets} \\
\midrule
CroSSL~\cite{deldari2024crossl} & latent masking & partial & 1--2 \\
LIMU-BERT~\cite{xu2021limu} & N/A (single modality) & small & 1 \\
Yuan et al.~\cite{yuan2024self} & N/A (single modality) & no & 1 (very large) \\
MultiMAE~\cite{bachmann2022multimae} & masking + reconstruction & no & 1 (vision) \\
PhysioME~\cite{lee2025physiome} & restoration decoder & no & 2 \\
PhysioOmni~\cite{jiang2025towards} & decoupled codebooks & no & 6 \\
CognifySSL v2.0~\cite{cisse2026detect} & presence predictor + decoder & no & 2 \\
\textbf{HeteroWear (ours)} & \textbf{padding mask, no reconstruction} & \textbf{yes} & \textbf{8} \\
\bottomrule
\end{tabular}

\vspace{4pt}
\parbox{\textwidth}{\small ``Cross-device'' indicates whether the pretraining corpus spans multiple hardware platforms for the same modality (e.g., phone- and watch-worn accelerometers), the heterogeneity targeted by HeteroWear.}
\end{table*}

Table~\ref{tab:related-comparison} summarizes where HeteroWear sits relative to the closest prior systems. PhysioME, PhysioOmni, and CognifySSL v2.0 share a repair philosophy for missing modalities and evaluate on modality sets (e.g., EEG/EOG/ECG, or IMU/ECG/EDA) that are fixed across training and test. To our knowledge, none pretrains jointly across independently collected datasets with different device hardware for the \emph{same} modality, which is the specific heterogeneity that HHAR and WISDM present and that HeteroWear is built around. HeteroWear's padding-mask design has no reconstruction component, and correspondingly cannot produce an explicit imputed estimate of what a missing sensor would have read, only a joint representation that remains usable without it. Neither design is strictly better; they answer different questions.

\section{Datasets}
\label{sec:data}

HeteroWear pretrains on eight datasets, summarized in Table~\ref{tab:datasets}. PAMAP2~\cite{reiss2012introducing} records accelerometer, gyroscope, and magnetometer data from hand, chest, and ankle units across nine subjects performing eighteen structured activities spanning everyday movement (walking, sitting, ironing) and athletic exertion (running, cycling, rope jumping). WISDM~\cite{weiss2019smartphone} records phone- and watch-worn accelerometer and gyroscope data from 51 subjects across 18 activities, including several fine hand-motion classes (eating soup, eating chips, eating a sandwich, folding clothes, typing) alongside coarser locomotion; it is the clearest cross-device instance in the corpus since the same activity is captured by two independent hardware platforms simultaneously. HHAR~\cite{stisen2015smart} was built specifically to study sensing heterogeneity across a mix of smartphone and smartwatch models with different sampling rates, covering six coarse locomotion activities. HARTH~\cite{logacjov2021harth} records two chest- and thigh-worn accelerometers from 22 subjects in a free-living, professionally annotated setting rather than a controlled laboratory protocol. UCI-HAR~\cite{anguita2013public} is the most homogeneous, single-device dataset in the corpus, phone-worn accelerometer and gyroscope only, with six basic locomotion classes. PPG-DaLiA~\cite{reiss2019deep} and WESAD~\cite{schmidt2018introducing} share an identical wrist-and-chest sensor pair; WESAD's labels are affect states rather than activities and contribute to pretraining only. MMASH~\cite{rossi2020public} is a single 24-hour free-living session per subject and likewise pretraining-only. Five of eight datasets provide usable activity labels for downstream evaluation, and HHAR downstream evaluation uses wrist-worn channels, following established convention~\cite{stisen2015smart}.

\begin{table}[htbp]
\caption{Pretraining corpus.\vspace{-0.4cm}}
\label{tab:datasets}
\begin{center}
\begin{tabular}{lrrl}
\toprule
\textbf{Dataset} & \textbf{Subj.} & \textbf{Windows} & \textbf{Modalities} \\
\midrule
PAMAP2$^*$   & 9  & 14{,}322 & ACC, GYRO, MAG, HR \\
WESAD        & 15 & 43{,}363 & ACC, ECG, EDA, RESP, TEMP \\
PPG-DaLiA    & 15 & 64{,}672 & PPG, ACC, ECG, RESP, TEMP \\
MMASH        & 22 & 58{,}027 & ACC, HR \\
WISDM$^*$    & 51 & 72{,}928 & ACC, GYRO \\
HHAR$^*$     & 9  & 44{,}212 & ACC, GYRO \\
HARTH$^*$    & 22 & 51{,}656 & ACC \\
UCI-HAR$^*$  & 30 & 9{,}081  & ACC, GYRO \\
\midrule
\textbf{Total} & \textbf{173} & \textbf{358{,}261} & \\
\bottomrule
\end{tabular}
\end{center}
\vspace{-0.1cm}\ \ \textit{Asterisked datasets have usable activity labels.}
\end{table}

Unifying eight independently collected datasets surfaced concrete data-quality problems that would otherwise have silently degraded pretraining. PAMAP2's raw inertial columns, not only its heart-rate column, encode dropped wireless packets as the literal string \texttt{NaN}; unfiltered, these silently poison every downstream per-window statistic. WISDM's raw timestamps are not globally monotonic within a subject, inflating apparent recording duration by orders of magnitude without correction. UCI-HAR records separate experiment sessions per subject without offsetting their local clocks, so naive windowing can select channels from two sessions that overlap in local time as though simultaneous; we deduplicate channels sharing modality and location, keeping the one with greater coverage. HHAR's raw device timestamps contain isolated multi-year jumps consistent with a hardware clock fault. Two of these were found via their downstream effect rather than manual inspection: the PAMAP2 corruption first surfaced as a masked-modeling loss that never left a four-digit value, and the value of per-instance signal normalization (Section~\ref{sec:method}) was established the same way, on a loss that converged quickly but plateaued roughly four orders of magnitude too high.

\section{Method}
\label{sec:method}

\begin{figure*}[t]
\centering
\includegraphics[width=\textwidth]{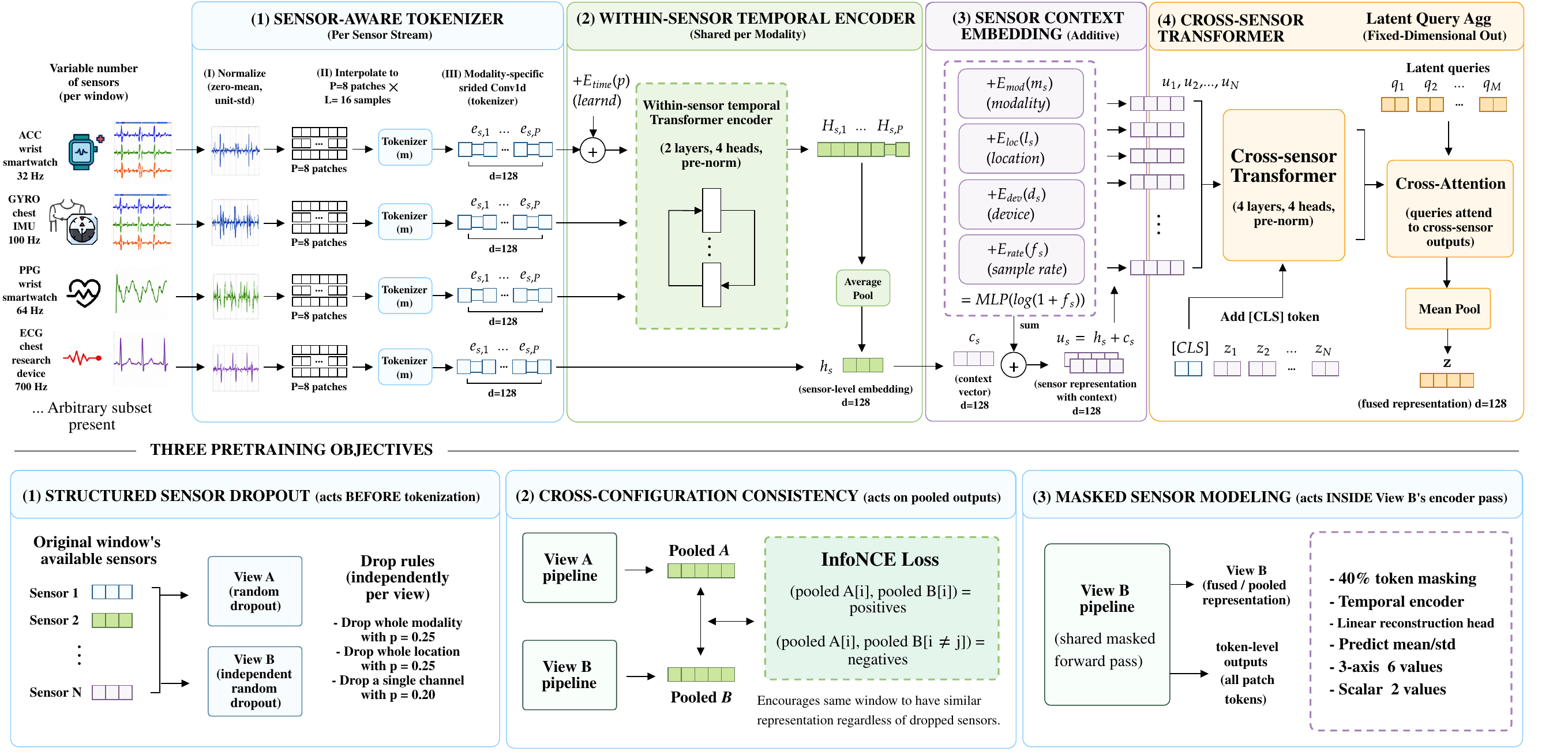}
\caption{\textbf{HeteroWear architecture.}
Sensor streams are temporally encoded, augmented with sensor-context
information, jointly modeled across sensors, and fused through learned
latent queries. The self-supervised objectives promote robustness to
heterogeneous and missing sensor configurations.}
\label{fig:architecture}
\end{figure*}

HeteroWear is designed around two distinct sources of structure in heterogeneous wearable data: temporal structure within an individual sensor stream and relational structure between simultaneously available sensors\footnote{Code: \url{https://github.com/yasamanmohammdpour/HeteroWear}}. Rather than concatenating every sensor patch into one global sequence, the architecture models these two levels hierarchically. Figure~\ref{fig:architecture} summarizes the complete architecture.

For each available sensor, HeteroWear first encodes its temporal patches with a modality-shared temporal Transformer. The resulting sensor-level representations are augmented with embeddings describing modality, body location, device class, and native sampling rate and then processed jointly by a cross-sensor Transformer. Finally, a small set of learned latent queries cross-attends to the available sensor representations to produce a fixed-dimensional fused representation for downstream prediction. This separation allows the same model to operate on arbitrary subsets of available sensors while assigning different modeling roles to temporal dynamics and cross-sensor interactions.

\subsection{Sensor-Aware Tokenization}
\label{subsec:tokenization}

Each window is represented as a variable-cardinality set of sensor streams. A sensor stream is a time series associated with one physical modality, body location, and device. Let
\begin{equation}
x_s \in \mathbb{R}^{T_s \times a_s}
\end{equation}
denote sensor $s$, where $a_s=3$ for vector-valued inertial modalities (accelerometer, gyroscope, and magnetometer) and $a_s=1$ for scalar physiological channels.

Each stream is normalized independently to zero mean and unit variance, preventing high-amplitude modalities from dominating the shared representation. Because datasets use different sampling rates and window durations, every present sensor is interpolated to $P=8$ temporal patches of $L=16$ samples, that is, 128 samples per stream regardless of its native rate. The native sampling rate is therefore not visible in the resampled signal and is supplied to the model separately through the sampling-rate embedding described below.

A modality-specific strided Conv1d tokenizer maps each patch to a $d=128$ representation,
\begin{equation}
e_{s,p} = \mathrm{Tokenizer}_{m(s)}\left(x_{s,p}\right), \qquad p=1,\ldots,P,
\end{equation}
where $m(s)$ denotes the modality of sensor $s$. Tokenizer parameters are shared across all datasets, devices, and body locations belonging to the same physical modality. Thus, for example, wrist, phone, chest, and ankle accelerometers use the same accelerometer tokenizer. A learned temporal-position embedding $E_{\mathrm{time}}(p)$ is added to each patch token,
\begin{equation}
\tilde e_{s,p} = e_{s,p}+E_{\mathrm{time}}(p).
\end{equation}
Fixing the number of patches rather than the raw number of samples provides a common temporal representation across heterogeneous native sampling rates while keeping the computational cost bounded.

\subsection{Within-Sensor Temporal Encoder}
\label{subsec:temporal-encoder}

Temporal structure is modeled independently for each available sensor before information is exchanged across sensors. For sensor $s$, its $P$ patch representations are processed by a two-layer, four-head pre-norm Transformer encoder,
\begin{equation}
H_s = \mathrm{Transformer}_{\mathrm{temp},m(s)}\left([\tilde e_{s,1},\ldots,\tilde e_{s,P}]\right).
\end{equation}
Temporal-encoder parameters are shared between sensors of the same modality, so an accelerometer observed on a wrist, phone, chest, or ankle is modeled with the same temporal dynamics while its provenance is represented separately through sensor context. The contextual patch representations are pooled to obtain one sensor-level embedding,
\begin{equation}
h_s = \frac{1}{P}\sum_{p=1}^{P} H_{s,p}.
\end{equation}
This stage deliberately isolates intra-sensor temporal modeling from cross-sensor fusion: patches from unrelated sensors do not compete for attention during temporal encoding.

\subsection{Sensor Context Embeddings}
\label{subsec:context}

The same modality can have different semantics depending on where and how it is measured. We therefore attach explicit provenance information to each sensor-level representation. For sensor $s$, the context vector is
\begin{equation}
c_s = E_{\mathrm{mod}}(m_s) + E_{\mathrm{loc}}(\ell_s) + E_{\mathrm{dev}}(d_s) + E_{\mathrm{rate}}(f_s),
\end{equation}
where modality, body location, and device class are represented by learned lookup embeddings. Native sampling rate $f_s$ is encoded using a small MLP,
\begin{equation}
E_{\mathrm{rate}}(f_s) = \mathrm{MLP}\left(\log(1+f_s)\right).
\end{equation}
The sensor representation passed to the cross-sensor stage is
\begin{equation}
u_s = h_s+c_s.
\end{equation}
Explicit context prevents the encoder from treating, for example, a research-grade 100~Hz chest accelerometer and a 32~Hz consumer wrist accelerometer as interchangeable simply because they measure the same physical modality.

\subsection{Cross-Sensor Relational Encoder}
\label{subsec:crosssensor}

Given the $S$ sensors available in a window, their contextualized representations are collected into
\begin{equation}
U = [u_1,\ldots,u_S] \in \mathbb{R}^{S\times d}.
\end{equation}
A four-layer, four-head cross-sensor Transformer then models relationships among available sensors,
\begin{equation}
V = \mathrm{Transformer}_{\mathrm{sensor}}(U).
\end{equation}
This stage learns relationships that are difficult to represent using independent sensor encoders alone, including redundancy between phone and watch accelerometers, complementary motion information across body locations, and interactions between physiological and inertial signals.

Critically, $S$ is not fixed. If a sensor is unavailable, its representation is simply absent from $U$. Batches are padded to the largest sensor set in the batch, and a padding mask prevents attention to nonexistent sensors. No missing sensor value or latent representation is imputed.

\subsection{Latent Cross-Attention Fusion}
\label{subsec:latentfusion}

A variable sensor set must ultimately be converted to a fixed-dimensional representation for downstream prediction. Rather than mean-pooling all sensors into one vector, we use $K=4$ learned latent queries
\begin{equation}
Q\in\mathbb{R}^{K\times d}.
\end{equation}
The latent queries cross-attend to the contextualized sensor representations, with the available sensor representations providing keys and values,
\begin{equation}
Z = \mathrm{CrossAttn}(Q,V,V).
\end{equation}
A subsequent latent self-attention block allows information captured by different queries to interact,
\begin{equation}
\tilde Z = \mathrm{Transformer}_{\mathrm{latent}}(Z).
\end{equation}
The final window-level representation is
\begin{equation}
g = \frac{1}{K}\sum_{k=1}^{K}\tilde Z_k,
\end{equation}
and the downstream classification head operates on $g$. This latent-fusion stage serves two purposes. First, its computational cost depends only weakly on the number of available sensors because the number of learned queries is fixed. Second, it allows the model to selectively aggregate complementary or redundant sensor information rather than requiring a single pooled vector to summarize every sensor interaction.

\subsection{Self-Supervised Pretraining Objectives}
\label{subsec:objectives}

HeteroWear combines three complementary pretraining mechanisms. Structured sensor dropout changes which sensors are available to the model; cross-configuration consistency encourages different sensor subsets of the same underlying window to produce similar global representations; and masked sensor modeling requires each temporal encoder to capture local signal structure.

\subsubsection{Structured Sensor Dropout}

For every augmented view, complete modalities are independently removed with probability $p_{\mathrm{mod}}=0.25$, complete body locations with probability $p_{\mathrm{loc}}=0.25$, and remaining individual sensor channels with probability $p_{\mathrm{ch}}=0.20$. If all sensors would otherwise be removed, one sensor is restored at random. Dropout is applied \emph{before} the temporal encoders, so removed sensors are genuinely absent from both the temporal and cross-sensor computation. The hierarchy approximates three realistic failure modes: loss of an entire sensor type, disappearance of a complete worn device or body location, and failure of an individual channel while colocated sensors remain available.

\subsubsection{Cross-Configuration Consistency}

Two independently sampled structured-dropout views of the same window, $x_i^{(A)}$ and $x_i^{(B)}$, are processed through the complete HeteroWear encoder and latent-fusion stage, producing fused representations $g_i^{(A)}$ and $g_i^{(B)}$. We optimize a symmetric InfoNCE objective~\cite{chen2020simple},
\begin{equation}
\mathcal{L}_{\mathrm{cons}}
=\tfrac12\!\left[
\mathrm{CE}(S_{AB}/\tau,y)
+\mathrm{CE}(S_{BA}/\tau,y)
\right],
\end{equation}
where $S_{AB}=\mathrm{sim}(G^A,G^B)$, $S_{BA}=\mathrm{sim}(G^B,G^A)$, $y_i=i$ identifies the matching view of sample $i$, and $\tau=0.1$. The objective encourages the fused representation of the same physical window to remain stable even when different modalities, locations, or individual channels are available.

\subsubsection{Per-Axis Masked Sensor Modeling}
\label{subsec:msm}

Masked sensor modeling (MSM) operates within the temporal stage. A fraction $r=0.4$ of real patch tokens from present sensors is replaced by a learned mask embedding before the temporal Transformer. For each masked patch, a linear prediction head applied to the corresponding contextual temporal output predicts the normalized per-axis mean and standard deviation of the original patch. Vector-valued inertial sensors therefore use six targets (mean and standard deviation for each of three axes), while scalar modalities use the corresponding scalar statistics.

The per-axis target preserves directional structure that is lost when all axes are pooled into one statistic. In a preliminary development experiment, replacing the pooled target with the per-axis formulation improved linear-probe macro-F1 by 0.043 on WISDM and 0.032 on HARTH. Unlike missing-modality reconstruction systems, MSM reconstructs only temporally masked patches from sensors that are actually present; a sensor removed by structured dropout has no token and no reconstruction target.

\subsection{Overall Pretraining Objective}

The complete objective is
\begin{equation}
\mathcal{L} = \mathcal{L}_{\mathrm{cons}} + \lambda\mathcal{L}_{\mathrm{msm}}, \qquad \lambda=1.
\end{equation}
Structured sensor dropout does not introduce a separate scalar loss; instead, it defines the sensor configurations supplied to both the consistency and masked-modeling objectives. The complete corpus contains eight datasets and 358{,}261 windows, but the primary controlled experiments use the leave-one-dataset-out protocol of Section~\ref{sec:setup}: when dataset $D$ is evaluated downstream, all windows from $D$ are excluded from self-supervised pretraining and the encoder is trained only on the remaining seven datasets. Unless otherwise specified, pretraining uses AdamW, batch size 64, a 400-step linear learning-rate warmup to $3\times10^{-4}$ followed by cosine decay, and gradient clipping at 1.0.

\section{Experimental Setup}
\label{sec:setup}

\subsection{Baselines and Controls}
\label{subsec:baselines}

We evaluate against four supervised HAR baselines trained from scratch per dataset on a single primary accelerometer channel: a random forest (RF) using 15 handcrafted statistics, a three-block one-dimensional CNN, DeepConvLSTM (DCLSTM)~\cite{ordonez2016deep}, and a compact three-layer patch Transformer (Transformer). These baselines follow the conventional single-stream HAR setting and use exactly the same downstream subject splits and window boundaries as HeteroWear. They are our own implementations under our protocol, with a single accelerometer channel as input, whereas HeteroWear uses all sensors available in each dataset.

To isolate the effect of self-supervised pretraining from architecture and sensor availability, we train an identical full-channel hierarchical model from random initialization using only downstream supervision (\textsc{HW-Scratch}). \textsc{HW-LODO} is the same architecture with heterogeneous leave-one-dataset-out pretraining. Consequently, \textsc{HW-Scratch} and \textsc{HW-LODO} differ only in whether heterogeneous self-supervised pretraining is used. \textsc{HW-Scratch} is not trained with structured sensor dropout.

\subsection{Metric}
\label{subsec:metric}

Macro-averaged F1 is the primary metric throughout because activity-class distributions are imbalanced within every downstream dataset. Accuracy can overweight frequent classes and obscure failures on less represented activities, whereas macro-F1 weights each activity class equally.

\subsection{Leave-One-Dataset-Out Pretraining}
\label{subsec:lodo}

The primary controlled evaluation uses leave-one-dataset-out (LODO) self-supervised pretraining. For a downstream target dataset $D$, every window belonging to $D$ is excluded from pretraining. HeteroWear is pretrained using only the remaining seven datasets and is subsequently linear-probed or fine-tuned using labeled training subjects from $D$. Thus, when HHAR is the downstream benchmark, no labeled or unlabeled HHAR window is encountered during self-supervised representation learning. The same procedure is independently repeated for PAMAP2, WISDM, HARTH, and UCI-HAR. This protocol distinguishes genuine cross-dataset transfer from transductive pretraining on the eventual target distribution. LODO removes the target dataset but not its device types: other pretraining datasets contain similar hardware, such as the phones and watches in WISDM and HHAR.

\subsection{Subject-Level Evaluation}
\label{subsec:subject-eval}

All downstream evaluation is subject-disjoint. PAMAP2 and HHAR use exhaustive nine-fold leave-one-subject-out evaluation. For each fold, one subject is held out for testing and no window from that subject is used for training or validation. WISDM, HARTH, and UCI-HAR use subject-disjoint splits, and every method in a comparison uses the same splits and window boundaries. For every comparison between \textsc{HW-Scratch} and \textsc{HW-LODO}, the exact same subject folds, validation subjects, classifier hyperparameters, and stopping criteria are used. Model selection is based exclusively on validation macro-F1.

\subsection{Linear Probing and Fine-Tuning}

Linear probing freezes the pretrained encoder, extracts one representation per window with no masking or sensor dropout, and trains logistic regression on the resulting embeddings. Fine-tuning uses a two-stage schedule. During the first 15\% of optimization steps, the encoder remains frozen while only the classification head is trained. The encoder is subsequently unfrozen and jointly optimized using a cosine-decayed learning rate of $5\times10^{-5}$ for the encoder and $10^{-3}$ for the classification head. The best validation checkpoint is used for test evaluation.

\subsection{Single-Channel SSL Control}
\label{subsec:single-ssl}

To determine whether generic masked self-supervised pretraining alone explains HeteroWear's performance, the compact Transformer used as the supervised baseline is pretrained jointly on the primary channels of the five downstream datasets using per-axis masked sensor modeling, then fine-tuned independently on each dataset. This model is denoted Transformer$_{\mathrm{SSL}}$.

\subsection{Missing-Sensor Evaluation}
\label{subsec:missing-eval}

Sensor robustness is evaluated after downstream training, without retraining the classifier on degraded inputs. A fixed classifier is first trained using full-channel representations. At test time, a fraction $d\in\{0,0.25,0.50,0.75\}$ of available channels is removed while always retaining at least one channel. The controlled comparison applies exactly the same randomly sampled 50\%-channel-removal masks to \textsc{HW-Scratch} and \textsc{HW-LODO}. We define retention as
\begin{equation}
R_{50} = \frac{F1_{\mathrm{50\%\ channel\ loss}}}{F1_{\mathrm{full}}}\times 100.
\end{equation}

\subsection{Held-Out-Device Evaluation}
\label{subsec:device-eval}

Because subject-level evaluation does not directly establish generalization to unseen hardware, we additionally evaluate device-level transfer on WISDM and HHAR. For WISDM, phone-to-watch and watch-to-phone transfer are evaluated separately. For HHAR, complete device models are held out from downstream training and used exclusively for testing. Both \textsc{HW-Scratch} and \textsc{HW-LODO} use identical device splits, making the comparison a direct test of whether heterogeneous pretraining improves transfer to unseen sensing hardware.

\subsection{Mechanism Ablation}
\label{subsec:mechanism}

The full objective combines masked sensor modeling (MSM), cross-configuration consistency, and structured sensor dropout. We run the ablation on HHAR because it is the benchmark on which heterogeneous-device transfer produces the clearest controlled improvement. Under the LODO protocol and the hierarchical architecture, we retrain with (i) MSM only, (ii) MSM plus cross-configuration consistency without structured dropout, (iii) MSM plus structured dropout without consistency, and (iv) the full objective. Separately, we remove the sensor-context embeddings while keeping the full objective and architecture. All variants use the same downstream folds and training procedure, and we report full-channel macro-F1, macro-F1 under 50\% channel removal, and retention.

\section{Results}
\label{sec:results}

\subsection{Controlled Cross-Dataset Transfer}
\label{subsec:controlled-transfer}

\begin{table*}[t]
\caption{Subject-level macro-F1 across five downstream datasets. RF, CNN, DCLSTM, and Transformer use a single accelerometer channel; Scratch and LODO use all available sensors. $\Delta_{\mathrm{SSL}}$ = LODO $-$ Scratch.}
\label{tab:controlled-transfer}
\centering
\small
\setlength{\tabcolsep}{9pt}
\renewcommand{\arraystretch}{1.05}
\begin{tabular}{lccccccc}
\toprule
\textbf{Dataset} & \textbf{RF} & \textbf{CNN} & \textbf{DCLSTM} & \textbf{Transformer} & \textbf{HW-Scratch} & \textbf{HW-LODO} & $\boldsymbol{\Delta_{\mathrm{SSL}}}$ \\
\midrule
PAMAP2  & .609 & .733 & .475 & .748 & .731 & \textbf{.756} & \textbf{+.025} \\
WISDM   & .627 & .716 & .686 & \textbf{.718} & .676 & .704 & \textbf{+.028} \\
HHAR    & .532 & .513 & .428 & .563 & .741 & \textbf{.823} & \textbf{+.082} \\
HARTH   & .431 & .498 & .470 & .547 & .526 & \textbf{.552} & \textbf{+.026} \\
UCI-HAR & .660 & .775 & .769 & .779 & .764 & \textbf{.786} & \textbf{+.022} \\
\midrule
\textbf{Mean} & .572 & .647 & .566 & .671 & .688 & \textbf{.724} & \textbf{+.037} \\
\bottomrule
\end{tabular}
\end{table*}

Table~\ref{tab:controlled-transfer} provides the primary comparison. \textsc{HW-Scratch} and \textsc{HW-LODO} use the identical hierarchical architecture, complete available sensor set, subject folds, and downstream training procedure; their only systematic difference is heterogeneous self-supervised pretraining. Pretraining improves the identical architecture on all five downstream benchmarks. Mean macro-F1 rises from 0.688 for HW-Scratch to 0.724 for HW-LODO, an average absolute improvement of 0.037. The gains are comparatively consistent on PAMAP2 (+0.025), WISDM (+0.028), HARTH (+0.026), and UCI-HAR (+0.022), while HHAR exhibits a substantially larger +0.082 improvement.

HW-Scratch already improves HHAR from 0.563 (single-accelerometer Transformer) to 0.741, showing that access to multiple sensor streams and the hierarchical architecture account for part of the gap. Self-supervised pretraining then raises performance from 0.741 to 0.823. HHAR is entirely absent from LODO pretraining, so this improvement does not depend on prior exposure to unlabeled HHAR windows.

Relative to the single-accelerometer Transformer, HW-LODO obtains the highest macro-F1 on four of five datasets, with modest margins on PAMAP2 (+0.008), HARTH (+0.005), and UCI-HAR (+0.007) and a large one on HHAR (+0.260). WISDM is the exception, where the Transformer is better by 0.014. Because these baselines use one accelerometer channel while HeteroWear uses all available sensors, the primary comparison in Table~\ref{tab:controlled-transfer} is Scratch versus LODO.

\subsection{Held-Out-Device Generalization}
\label{subsec:device-results}

\begin{table}[t]
\caption{Held-out-device macro-F1. HW-Scratch and HeteroWear use identical target-device splits.}
\label{tab:device-transfer}
\centering
\small
\setlength{\tabcolsep}{3.5pt}
\renewcommand{\arraystretch}{1.08}
\begin{tabular}{llccc}
\toprule
\textbf{Dataset} & \textbf{Transfer} & \textbf{Scratch} & \textbf{HeteroWear} & $\boldsymbol{\Delta}$ \\
\midrule
WISDM & phone $\rightarrow$ watch & 0.491 & \textbf{0.556} & \textbf{+0.065} \\
WISDM & watch $\rightarrow$ phone & 0.538 & \textbf{0.591} & \textbf{+0.053} \\
HHAR  & held-out device A         & 0.676 & \textbf{0.755} & \textbf{+0.079} \\
HHAR  & held-out device B         & 0.703 & \textbf{0.771} & \textbf{+0.068} \\
\midrule
\textbf{Mean} & --- & 0.602 & \textbf{0.668} & \textbf{+0.066} \\
\bottomrule
\end{tabular}
\end{table}

Table~\ref{tab:device-transfer} tests the cross-device hypothesis directly by evaluating hardware configurations excluded from downstream training. HeteroWear improves macro-F1 in every held-out-device condition, with absolute gains from +0.053 to +0.079 and a mean improvement of +0.066. On WISDM, phone-to-watch transfer improves from 0.491 to 0.556 and watch-to-phone from 0.538 to 0.591; on HHAR, the two held-out-device evaluations improve by +0.079 and +0.068.

This clarifies a distinction that subject-level evaluation cannot establish. The advantage is not limited to recognizing unseen participants on familiar devices; it persists when the downstream hardware changes. Together, LODO pretraining and held-out-device evaluation test two forms of generalization: transfer to an unseen dataset during representation learning, and transfer to unseen hardware during downstream evaluation.

\subsection{Controlled Missing-Sensor Robustness}
\label{subsec:controlled-robustness}

We next compare HW-Scratch and HW-LODO under identical test-time sensor failures. Table~\ref{tab:controlled-robustness} reports full-channel macro-F1, performance with 50\% sensor removal, and relative retention.

\begin{table*}[t]
\caption{Robustness under 50\% test-time sensor removal. S: HW-Scratch; L: HW-LODO; Ret.: $F1_{50}/F1_{\mathrm{full}}$.}
\label{tab:controlled-robustness}
\centering
\small
\setlength{\tabcolsep}{9pt}
\renewcommand{\arraystretch}{1.05}
\begin{tabular}{lcccccc}
\toprule
\textbf{Dataset} & \textbf{S-Full} & \textbf{S-50} & \textbf{S-Ret.} & \textbf{L-Full} & \textbf{L-50} & \textbf{L-Ret.} \\
\midrule
PAMAP2  & .731 & .402 & 55.0\% & .756 & .498 & \textbf{65.9\%} \\
WISDM   & .676 & .447 & 66.1\% & .704 & .526 & \textbf{74.7\%} \\
HHAR    & .741 & .563 & 76.0\% & .823 & .691 & \textbf{84.0\%} \\
HARTH   & .526 & .336 & 63.8\% & .552 & .409 & \textbf{74.1\%} \\
UCI-HAR & .764 & .456 & 59.7\% & .786 & .553 & \textbf{70.4\%} \\
\midrule
\textbf{Mean} & -- & -- & 64.1\% & -- & -- & \textbf{73.8\%} \\
\bottomrule
\end{tabular}
\end{table*}

\begin{figure*}[t]
\centering
\includegraphics[width=0.85\textwidth]{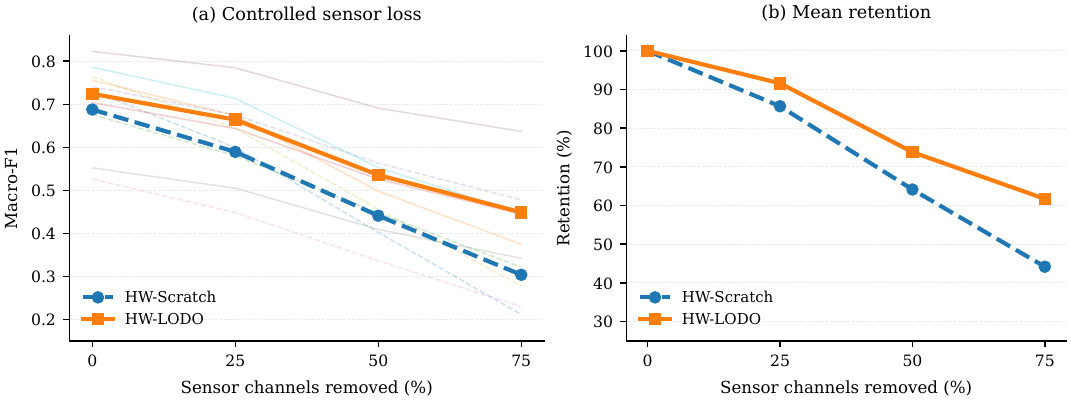}
\caption{\textbf{Robustness to sensor loss.}
Macro-F1 and mean performance retention for HW-Scratch and HW-LODO
under increasing test-time sensor removal.}
\label{fig:sensor-loss}
\end{figure*}

HW-Scratch retains an average of 64.1\% of its full-channel macro-F1 after half of the available channels are removed, whereas HeteroWear retains 73.8\%, an improvement of 9.7 percentage points. The retention benefit is observed on every dataset: +10.9 points on PAMAP2, +8.6 on WISDM, +8.0 on HHAR, +10.3 on HARTH, and +10.7 on UCI-HAR. Robustness does not simply track the full-channel advantage: HARTH improves by only +0.026 macro-F1 under full sensor availability, yet its retention increases from 63.8\% to 74.1\%. Figure~\ref{fig:sensor-loss} shows that HW-LODO degrades more gracefully than HW-Scratch as sensor availability decreases. Since HW-Scratch is not trained with structured sensor dropout, the retention gap reflects pretraining with structured dropout; the mechanism ablation below isolates which pretraining components drive it.

\subsection{Mechanism Ablation}
\label{subsec:mechanism-results}

To determine which components produce the HHAR transfer and robustness advantage, Table~\ref{tab:hhar-ablation} removes the principal pretraining mechanisms while preserving the hierarchical architecture and LODO protocol.

\begin{table}[t]
\caption{Mechanism ablation on HHAR. Retention is $F1_{50}/F1_{\mathrm{full}}$.}
\label{tab:hhar-ablation}
\centering
\small
\setlength{\tabcolsep}{4pt}
\renewcommand{\arraystretch}{1.08}
\begin{tabular}{lccc}
\toprule
\textbf{Configuration} & \textbf{F1} & \textbf{F1@50\%} & \textbf{Ret.} \\
\midrule
Scratch (no pretraining) & 0.741 & 0.563 & 76.0\% \\
MSM only & 0.758 & 0.560 & 73.9\% \\
MSM + consistency & 0.787 & 0.606 & 77.0\% \\
MSM + struct.\ dropout & 0.801 & 0.650 & 81.1\% \\
\textbf{Full objective} & \textbf{0.823} & \textbf{0.691} & \textbf{84.0\%} \\
Full, no context emb. & 0.781 & 0.639 & 81.8\% \\
\bottomrule
\end{tabular}
\end{table}

MSM alone increases full-channel macro-F1 from 0.741 to 0.758 but does not improve sensor-loss robustness: retention decreases slightly from 76.0\% to 73.9\%. Masked reconstruction therefore provides useful representation learning but is insufficient to explain graceful degradation under changing sensor configurations. Since this variant is pretrained on the same heterogeneous corpus, it also indicates that heterogeneous pretraining alone does not account for the retention gain.

Adding cross-configuration consistency without structured dropout raises macro-F1 to 0.787 with 77.0\% retention. Structured dropout without consistency reaches 0.801 macro-F1 and 81.1\% retention, indicating that explicit exposure to changing sensor availability has the larger effect on robustness. The complete objective performs best on both criteria, reaching 0.823 macro-F1 and 0.691 after 50\% channel loss (84.0\% retention). Relative to MSM alone, the full objective improves macro-F1 by 0.065 and retention by 10.1 percentage points. The mechanisms therefore appear complementary: consistency encourages representations of the same underlying window to remain similar across sensor configurations, while structured dropout exposes the encoder to the failure patterns it later encounters at inference time.

Removing the modality, location, device, and rate context decreases macro-F1 from 0.823 to 0.781 despite preserving the hierarchical architecture and complete SSL objective. Retention remains comparatively high (81.8\%), so sensor context appears to contribute mainly to accuracy rather than robustness. This supports explicitly identifying sensor provenance when a shared representation is learned across heterogeneous devices and body locations.

\subsection{Single-Channel Self-Supervised Control}
\label{subsec:single-ssl-results}

Table~\ref{tab:ssl-ablation} asks whether masked reconstruction alone improves the conventional Transformer when heterogeneous multimodal modeling is removed.

\begin{table}[t]
\caption{Single-channel masked-reconstruction SSL control (macro-F1).}
\label{tab:ssl-ablation}
\centering
\small
\setlength{\tabcolsep}{5pt}
\begin{tabular}{lccc}
\toprule
\textbf{Dataset} & \textbf{Transf.} & \textbf{Transf.$_{\mathrm{SSL}}$} & $\boldsymbol{\Delta}$ \\
\midrule
PAMAP2  & \textbf{0.748} & 0.671 & $-0.077$ \\
WISDM   & \textbf{0.718} & 0.512 & $-0.206$ \\
HHAR    & 0.563 & \textbf{0.634} & $+0.071$ \\
HARTH   & \textbf{0.547} & 0.253 & $-0.294$ \\
UCI-HAR & \textbf{0.779} & 0.706 & $-0.073$ \\
\bottomrule
\end{tabular}
\end{table}

Single-channel masked reconstruction does not reproduce the behavior of HeteroWear. It decreases downstream macro-F1 on PAMAP2, WISDM, HARTH, and UCI-HAR and improves only HHAR. Read together with Table~\ref{tab:controlled-transfer}, where the hierarchical HeteroWear model improves over its matched scratch control on all five datasets, this suggests that the gain depends on the interaction between heterogeneous sensor modeling and the pretraining objectives rather than on self-supervision in isolation.

\section{Discussion}
\label{sec:discussion}

The controlled hierarchical comparison separates the effects of self-supervised pretraining from those of architecture and richer sensor input. HW-Scratch and HW-LODO share the same architecture and sensor inputs, yet LODO pretraining improves mean macro-F1 from 0.688 to 0.724 across all five datasets, with the largest gain on HHAR (+0.082), showing that heterogeneous pretraining provides benefits beyond simply exposing the classifier to more sensors. Held-out-device experiments are consistent with this: HeteroWear improves macro-F1 on held-out devices by 0.053--0.079 across WISDM and HHAR, with a mean gain of 0.066, indicating that the HHAR improvement is not due only to a favorable subject split and that the representation transfers when the downstream sensing hardware changes.

HeteroWear also retains more performance under missing sensors, keeping 73.8\% of full-channel performance after 50\% channel removal compared with 64.1\% for the identical architecture trained from scratch. The ablations show that MSM alone provides little robustness while structured sensor dropout substantially improves retention, which supports treating changing sensor availability as a pretraining condition rather than repairing missing channels only after deployment.

The advantage is not uniform across benchmarks. The single-accelerometer Transformer remains slightly stronger on WISDM (0.718 versus 0.704), while gains over it on PAMAP2, HARTH, and UCI-HAR are small, consistent with HeteroWear being designed primarily for heterogeneous and changing sensor configurations rather than for replacing task-specific supervised models on homogeneous HAR benchmarks. WISDM also shows that cross-device heterogeneity alone does not guarantee an advantage: compared with HHAR, it contains finer activity distinctions, including multiple hand-centric activities that may require high-resolution local motion information. We do not vary corpus size, so the contributions of heterogeneity and of scale are not separated here.

\section{Limitations and Future Work}
\label{sec:limitations}

Our supervised baselines use a single accelerometer channel and are our own implementations, so Table~\ref{tab:controlled-transfer} is best read as a controlled study of pretraining rather than a state-of-the-art comparison. HW-Scratch is not trained with sensor dropout, so the retention gap reflects pretraining with structured dropout rather than heterogeneous data alone. Results are single-run without significance tests, PAMAP2 and HHAR have nine subjects each, and LODO removes the target dataset but not its device types. We also do not ablate individual sensor types, sweep sampling rates, or report latency. Future work includes published self-supervised and multi-channel baselines under identical splits, a scratch control with sensor dropout, per-sensor and sampling-rate ablations, and profiling on wearable-class hardware with efficiency techniques such as precision-adaptive optimization~\cite{sheibanian2025tri}. Controlled synthetic benchmarks with adjustable heterogeneity, built in a modular fashion~\cite{shaeri2025mnist}, could separate heterogeneity from scale, and human-in-the-loop feedback~\cite{shaeri2025explainable} could help adapt a pretrained representation to a new device.

\section{Conclusion}
\label{sec:conclusion}

HeteroWear learns wearable representations across eight heterogeneous datasets by combining within-sensor temporal modeling, sensor-context encoding, cross-sensor attention, latent fusion, structured sensor dropout, cross-configuration consistency, and masked sensor modeling. Controlled evaluation separates the effect of pretraining from architecture and sensor availability. With the target dataset excluded from self-supervised pretraining, HeteroWear improves the identical full-channel hierarchical architecture trained from scratch on all five downstream benchmarks, increasing mean macro-F1 from 0.688 to 0.724, and it exceeds a single-accelerometer Transformer on four of five datasets, with the largest gain on heterogeneous-device HHAR. The advantage persists under hardware and sensor-availability shifts: HeteroWear improves held-out-device macro-F1 by 0.066 on average and retains 73.8\% of full-channel performance after 50\% channel removal, compared with 64.1\% for the identical architecture trained from scratch without sensor dropout. Ablations show that structured sensor dropout is particularly important for robustness, while cross-configuration consistency and sensor context further improve transfer. Overall, the results support explicitly modeling sensor identity and availability during representation learning.

\bibliographystyle{IEEEtran}
\bibliography{references}

\end{document}